\documentclass[10pt,twocolumn,letterpaper]{article}

\usepackage[pagenumbers]{cvpr} 

\definecolor{cvprblue}{rgb}{0.21,0.49,0.74}
\usepackage[pagebackref,breaklinks,colorlinks,allcolors=cvprblue]{hyperref}

\def\paperID{*****} 
\def\confName{CVPR}
\def\confYear{2025}

\title{Dietary Intake Estimation via Continuous 3D Reconstruction of Food}

\author{Wallace Lee\\
University of Waterloo \\
{\tt\small wwlee@uwaterloo.ca}
\and
YuHao Chen\\
University of Waterloo\\
{\tt\small yuhao.chen1@uwaterloo.ca}
}

\begin{document}
\maketitle
\begin{abstract}
Monitoring dietary habits is crucial for preventing health risks associated with overeating and undereating, including obesity, diabetes, and cardiovascular diseases. Traditional methods for tracking food intake rely on self-reported data before or after the eating, which are prone to inaccuracies. This study proposes an approach to accurately monitor ingest behaviours by leveraging 3D food models constructed from monocular 2D video. Using COLMAP and pose estimation algorithms, we generate detailed 3D representations of food, allowing us to observe changes in food volume as it is consumed. Experiments with toy models and real food items demonstrate the approach’s potential. Meanwhile, we have proposed a new methodology for automated state recognition challenges to accurately detect state changes and maintain model fidelity. The 3D reconstruction approach shows promise in capturing comprehensive dietary behaviour insights, ultimately contributing to the development of automated and accurate dietary monitoring tools.
\end{abstract}    
\section{Introduction}
\label{sec:intro}
\begin{figure*}[t]
    \centering
    \includegraphics[scale=0.45]{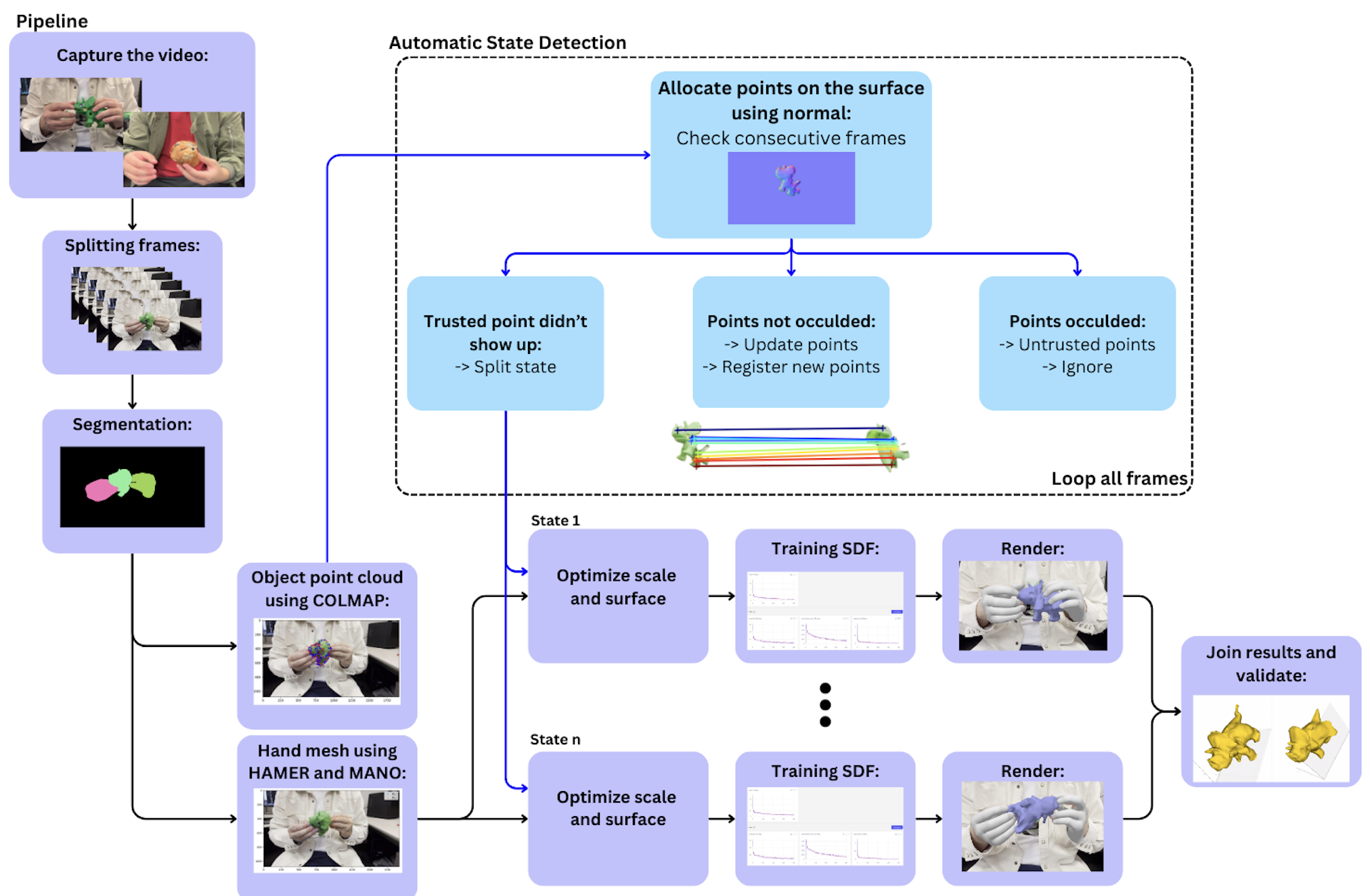}
    \caption{A high-level diagram of the 3D food construction process. The purple blocks represent the steps of the base HOLD pipeline \cite{fan2024hold}, while the blue blocks indicate the new components added for sequence separation. The process begins with splitting the captured video into individual frames and segmenting the frames. The object masks are passed into COLMAP to construct the object point cloud, while hand masks are processed by HAMER and refined with MANO to generate the hand meshes. During the point cloud construction, frames are converted to normal maps to detect changes in the original topology to split the original dataset into individual sequences. The HOLD optimizer takes the separated hand meshes and object point cloud as inputs to optimize the surface and scale based on their interaction. At last, the optimized meshes and point cloud is then inputed to train an SDF model, which generates the refined object's mesh.
    }
    \label{fig1}
\end{figure*}

Food is an essential part of daily life; however, unhealthy eating habits, including overeating or undereating, are strongly associated with an increased risk of chronic conditions such as diabetes, obesity, and cardiovascular disease \cite{de2017healthy}. Therefore, monitoring and evaluating dietary habits is crucial to promote a healthier lifestyle and prevent these conditions.

Recent advancements in computer vision and machine learning (ML) saw the development of various algorithms and devices proposed for automatic monitoring of ingest behaviours, as an attempt to replace the reliance on self-reported dietary reports because it is heavily dependent on user compliance. Current methods primarily focus on predicting when individuals consume food or beverages through human gesture prediction, such as jaw motion, hand gestures, and body position with the help of video analysis, wearable or remote sensors \cite{wang2022eating, tang2022new, fontana2014automatic}. While these methods can track the digestive behaviour of an individual based on counting bites and intervals, they fail to consider food-specific attributes, such as the quantity consumed or the macro-nutrient content. This limitation results in a partial assessment of an individual’s dietary habits. Hence, it highlights the need for methods that integrate food volume analysis into dietary monitoring systems for a more holistic understanding of eating behaviours. 

In response to the recent significant progress in the 3D reconstructing field, this paper proposes an approach to enhance dietary monitoring by leveraging 3D reconstructing methodologies to model food characteristics. The main objective is to capture 3D food models through a monocular 2D video with a structure-from-motion and multi-view-stereo pipeline that enables point cloud generation \cite{schoenberger2016mvs, schoenberger2016sfm}. This approach enables precise modeling of food items, which enables the consideration of both quantity and the changing topology of the food during consumption.  

To enhance the accuracy of the food object models, we build upon the HOLD model \cite{fan2024hold} for pose estimation which is a method to refine the output with hand interaction, due to the common appearance of hand-held food such as snacks, sandwiches, and wraps.  

Preliminary experiments revealed some challenges in maintaining accurate pose and point registration when parts of the object were detached which mimics bites in these experiments. This results in the rendered model being in the wrong shape, as it failed to distinguish the topological differences between the objects' intact and detached states and deemed it as a rigid deformation. Therefore we proposed a method to automatically identify the sudden change of shape during the point cloud generation phase of the pipeline such that model training could be done separately on distinct objects' forms or states as a recommended future work. 

The main takeaways of this study are the exploration of using 3D reconstruction of food models from video data to analyze ingest behaviours, as well as the proposed methodology to automatically detect structural changes $S_i$, thereby encapsulating a more comprehensive and dynamic assessment of dietary behaviours.

\section{Related Works}
\label{sec:related}

The Automatic Ingestion Monitor (AIM) introduced by Fontana et al. utilizes a wearable device to track ingestive behaviour by combining multiple sensors to monitor eating episodes non-invasively.  \cite{fontana2014automatic}. Similarly, other research groups explored radar-based monitoring using Frequency-Modulated Continuous Wave (FMCW) radar combined with a 3D temporal convolutional network for gesture detection \cite{wang2023eat}. These devices demonstrated high accuracy in detecting eating activities, with the latter offering substantial improvements in gesture prediction accuracy without requiring user attachment, addressing the discomfort often associated with wearable devices \cite{wang2023eat}.

Other research has also explored video-based methods for tracking food intake gestures. A dataset had been specifically designed for recognizing intake gestures in cafeteria settings, leveraging video data to track hand and jaw movements \cite{tang2022new}. At the same time, smartphones are also used for video monitoring to capture eating behaviours in home environments through deep learning models to recognize food intake \cite{wang2022eating}. These systems excel in natural settings, offering a more dynamic and accessible approach than sensor-based methods due to reduced setup requirements.

While all these methods provide substantial accuracy in tracking biting gestures and body movements for dietary monitoring, they fail to account for bite-size progression based on food geometry changes. This limitation hinders their ability to deliver a comprehensive view of dietary behaviour.

The HOLD pipeline, which serves as the foundation of this work, is a category-agnostic framework designed to reconstruct 3D hand-object interactions from monocular videos. One of its key capabilities lies in refining object poses by iteratively optimizing the alignment between the object’s 3D point cloud and its observed image features, ensuring accurate representation even in dynamic interaction scenarios. Additionally, HOLD effectively addresses hand occlusions by leveraging motion priors and hand-object interaction models to infer the occluded regions of both the hand and object. This allows it to reconstruct complete 3D meshes even when parts of the object are obscured by the hand. These advancements enable accurate modeling of diverse hand-held objects, ranging from everyday items like water bottles to complex geometric structures such as Lego models \cite{fan2024hold}. This functionality is particularly promising for dietary monitoring, given the prevalence of hand-held food items like sandwiches in daily life. By enabling real-time tracking of food topological changes, HOLD provides a dynamic and detailed assessment of food consumption, capturing the size of each bite and its impact on overall dietary behaviour.

\section{Methods}
\label{sec:methods}

The input for this methodology is the original RGB monoscopic 2D video, which aligns with HOLD. All data was recorded using toy objects and food items with an iPhone 14, mainly for consistency with the device used in the HOLD paper for in-the-wild sequences \cite{fan2024hold}. Each video sequence recorded is monocular, which captures the individual and their hands. The initial and final frames of each video were trimmed to exclude instances where the participant’s hands were out of frame when initiating or stopping the recording on the phone.

The preprocessing begins by splitting the video into individual frames, which are then automatically segmented to create hand masks and object masks \cite{cheng2023segment}. The hand masks are used to predict hand mesh using the pre-trained HAMER model \cite{pavlakos2024reconstructing}, which is further refined with the MANO hand model \cite{MANO:SIGGRAPHASIA:2017} for improved precision. Simultaneously, COLMAP is employed to estimate and predict the object’s pose and create the object's mesh base on the object's masks \cite{schoenberger2016mvs, schoenberger2016sfm}.

When both the 3D mesh models of the objects and hands are complete, they are optimized based on their interactions, with each object scaled to their relative sizes. A signed distance field (SDF) model is then trained to render the scene with accurately represented 3D objects and hands \cite{zheng2023locally}. The entire pipeline is summarized with Figure~\ref{fig1} 

In this experiment, some common handheld food objects along with a toy Triceratops with removable legs are used as test objects. The food would have a undergone a food volume reduction, while the removable toy is to mimic the same action with the extra benefit of complex geometry serving as a robust test case for the model’s capability in handling intricate shapes.

\begin{figure}[]
    \centering
    \includegraphics[scale=0.45]{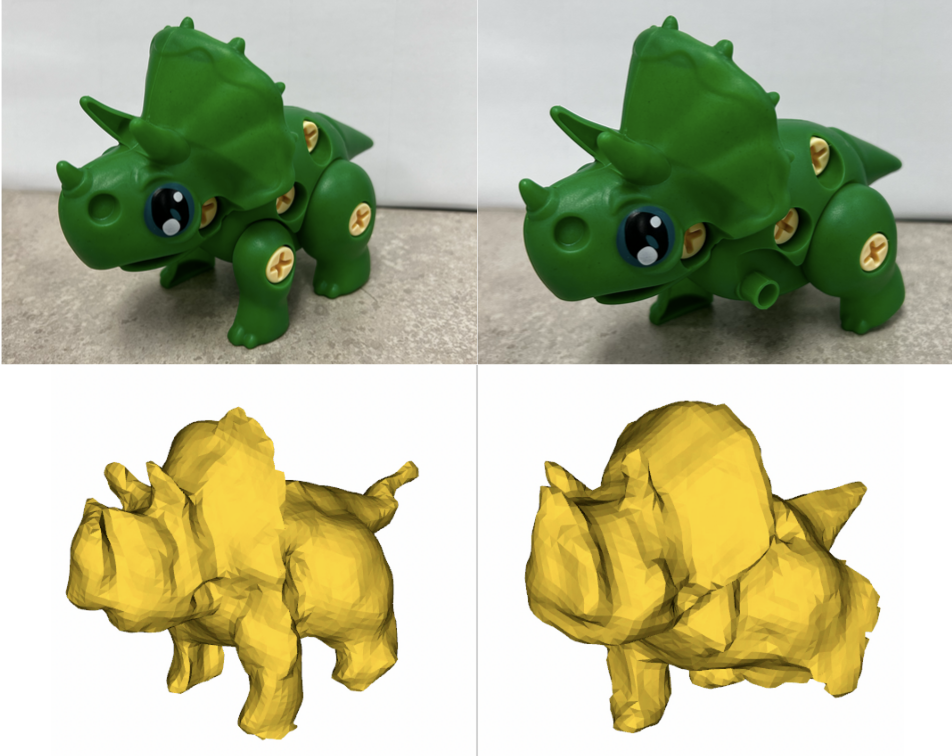}
    \caption{Top row: The actual toy triceratops experimented with | Bottom row: The 3D meshes of triceratops in with varied structures | Left: Before detachment of the leg | Right: After detachment of the leg}
    \label{fig2}
\end{figure}

To validate changes in the model over time, the reconstructed 3D models from each state $S_1 ... S_n$ are extracted, and the volume percentage decrease is measured. This calculated volume change is compared against ground truth measurements obtained using the water displacement method, providing a rough estimate of the model’s depiction of topology changes. Due to the porous nature of the Triceratops toy, an additional zip-lock bag is used for containing the toy, which was then vacuumed and submerged into the water container. an additional empty container is placed under to collect the overflow water, which is the indication of the volume displaced by the object. The zip-lock bag's volume is also collected in order to subtract out of the final measurement. The meshes are displayed in Figure~\ref{fig2}, and the volume calculation is done with the python library vedo \cite{musy2019vedo}.

\begin{figure}[]
    \centering
    \includegraphics[scale=0.375]{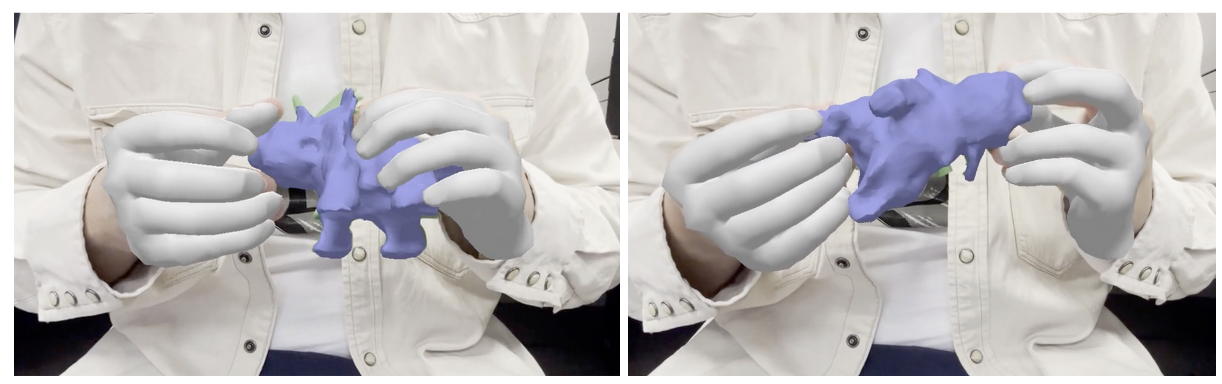}
    \caption{The rendered view of triceratops with hands in different topology. Purple colour mesh is the object and the hands' meshes are in white colour. Left: Before detachment | Right: After detachment}
    \label{fig3}
\end{figure}

\begin{figure}[]
    \centering
    \includegraphics[scale=0.4]{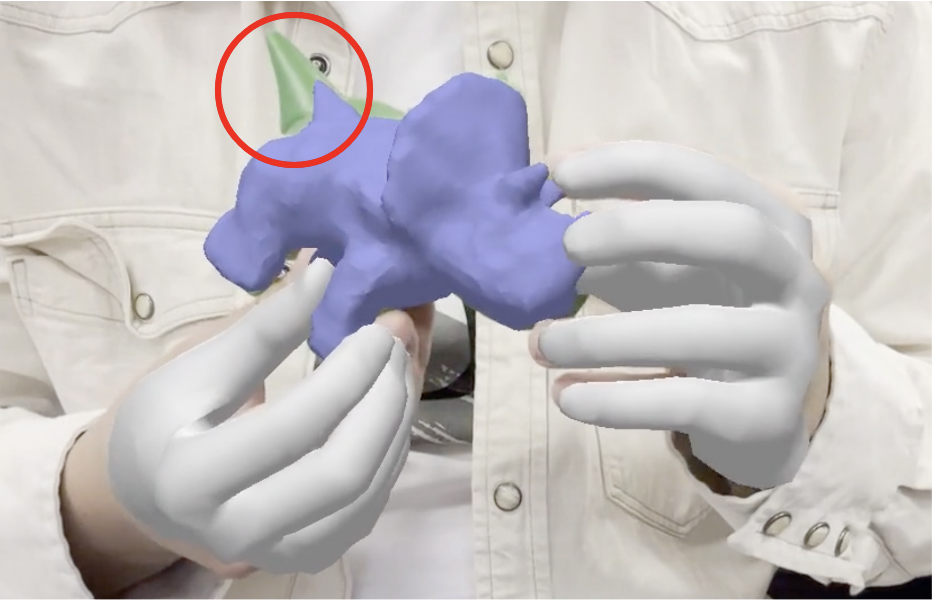}
    \caption{Failure case in the rendered view of triceratops, the red circle indicate the failing part}
    \label{fig4}
\end{figure}

\begin{figure}[]
    \centering
    \includegraphics[scale=0.25]{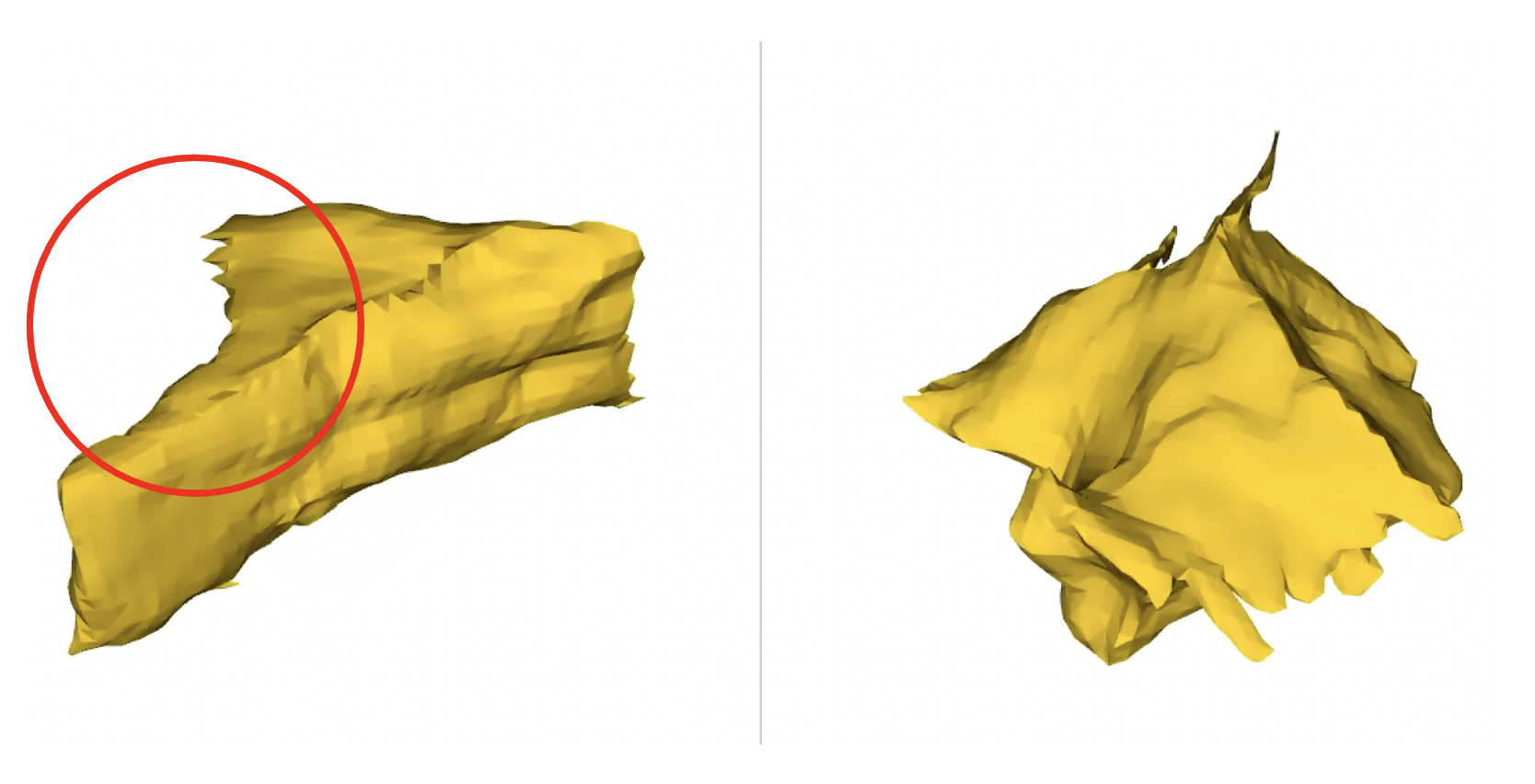}
    \caption{The 3D meshes of sandwich in with varied structures | Left: Before bite, the red circle highlight the missing part of the topology from the original sandwich object | Right: After bite}
    \label{fig5}
\end{figure}

\begin{figure}[]
    \centering
    \includegraphics[scale=0.3]{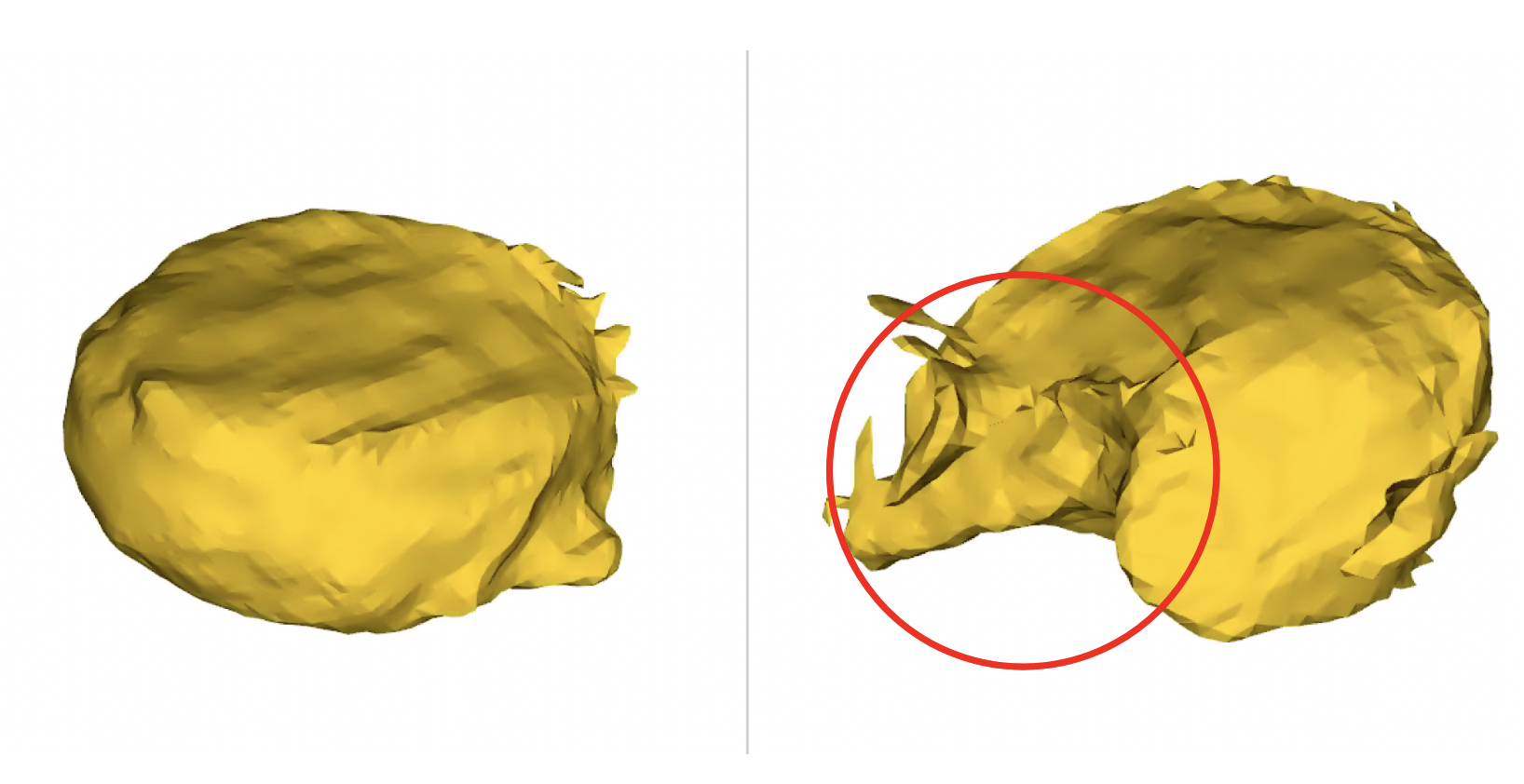}
    \caption{The 3D meshes of filled donut in with varied structures | Left: Before bite | Right: After bite, with the bitten area circled in red}
    \label{fig6}
\end{figure}

\begin{figure}[]
    \centering
    \includegraphics[scale=0.225]{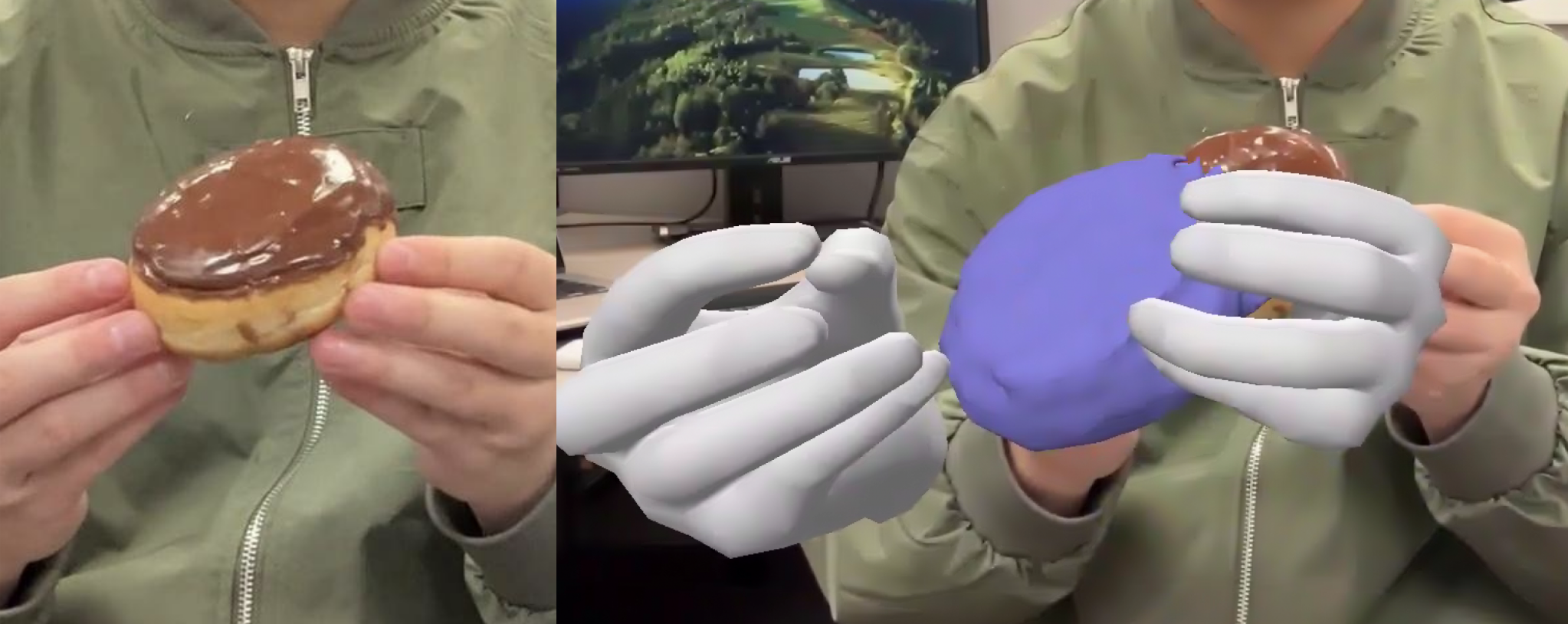}
    \caption{Left: Original RGB Image of the filled donut | Right: The rendered view of filled donut with hands. Purple colour mesh is the object and the hands' meshes are in white colour.}
    \label{fig7}
\end{figure}

\begin{figure}[]
    \centering
    \includegraphics[scale=0.3]{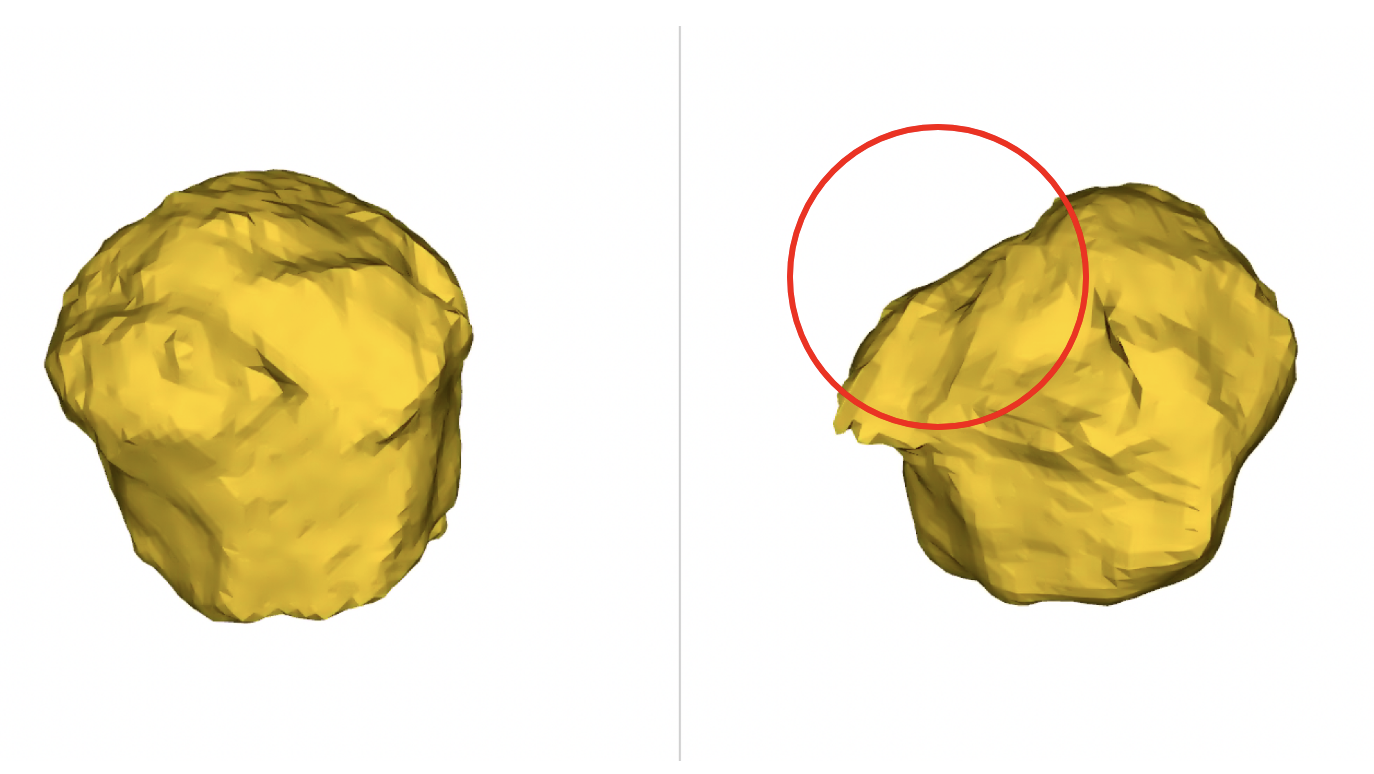}
    \caption{The 3D meshes of muffin in with varied structures | Left: Before bite | Right: After bite, with the bitten area circled in red}
    \label{fig8}
\end{figure}

\begin{figure}[]
    \centering
    \includegraphics[scale=0.3]{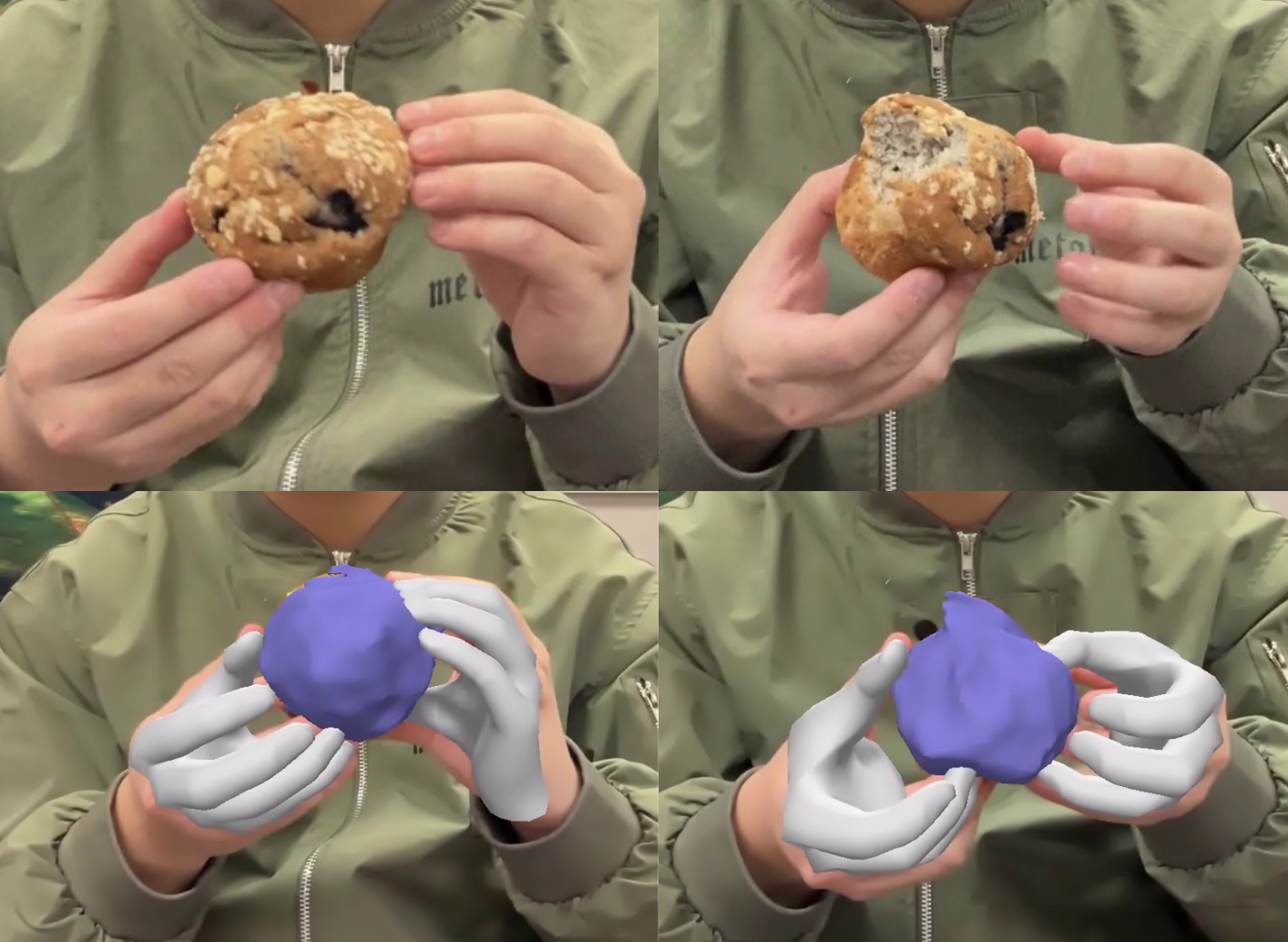}
    \caption{The muffin's original RGB image with their rendered view with hands in different topology. Purple colour mesh is the object and the hands' meshes are in white colour. Top left: Before bite original RGB | Top right: After bite original RGB | Bottom Left: Before bite rendered view | Bottom Right: After bite rendered view}
    \label{fig9}
\end{figure}
\section{Results}
\label{results}

\subsection{Qualitative}
The main qualitative results from Figure~\ref{fig2} to \ref{fig10} are derived from the rendered view from the hand and object's meshes aligned with the video, as well as the meshes themselves.

\subsection{Quantitative}
The main metric for comparison is volume percentage decrease as there is a lack of physical scale reference. Meanwhile, the percentage decrease would also serve as a useful indication of the relative quantity of food consumed from the video. 

\begin{table}[]
    \caption{Volume decrease difference between ground truth and 3D reconstructed meshes}
    \label{table1}
    \centering
    \resizebox{\columnwidth}{!}{%
        \begin{tabular}{cccc}
            \toprule
            \textbf{Object} & \textbf{Ground Truth} & \textbf{Prediction} & \multicolumn{1}{c}{\textbf{Difference}} \\
            \midrule
            Triceratops & 7.843\% & 6.595\% & -1.348\% \\
            Muffin & 30.43\% & 37.70\% & +7.27\% \\
            Sandwich & 58.33\% & 52.32\% & -6.01\% \\
            Filled Donut & 14.28\% & 20.88\% & +6.60\%\\
            \bottomrule
        \end{tabular}
    }

\end{table}

\section{Discussion}
\label{discussion}

\textbf{Qualitative Analysis:}
As shown in Figures~\ref{fig3} and~\ref{fig9}, the model qualitatively achieves strong alignment with the observed objects across most video frames, demonstrating consistent and stable results throughout the sequences.

In the triceratops example, there were isolated instances of synchronization issues, particularly noticeable with the tail failing to align accurately with the body's rotation as shown in Figure~\ref{fig4}. Despite these occasional misalignments, the overall mesh integrity remained stable across the sequences, suggesting that such discrepancies are minor and generally acceptable when considering the high alignment accuracy achieved in the majority of frames.

A closer inspection of the triceratops meshes before and after the leg detachment event (Figure~\ref{fig2}) highlights notable differences in resolution and completeness, especially in the tail region. Specifically, the mesh after leg detachment exhibited considerably higher surface detail, improved completeness, and a fully reconstructed tail, while the mesh before detachment appeared incomplete and had lower resolution. This disparity is primarily due to the varying number of frames used during the mesh generation process—267 frames for the initial mesh versus 452 frames post-detachment. Increasing the frame count directly enhances COLMAP’s capability to generate higher-fidelity point clouds, thus capturing finer geometric details more effectively \cite{schoenberger2016mvs, schoenberger2016sfm}. Furthermore, potential occlusion by the hands during recording may have partially obstructed the tail, further contributing to these observed differences.

In the food item reconstructions, the muffin exhibited better overall topological performance compared to the triceratops, benefiting from its simpler geometry. As clearly indicated by the red circle in Figure~\ref{fig8}, the bitten region was effectively identified and visualized in the reconstructed 3D mesh. However, minor surface distortions were still present, which can likely be attributed to inherent challenges of reconstructing relatively homogeneous textures—typical in foods like muffins. This homogeneity reduces the distinctiveness of visual features, complicating feature matching, and consequently resulting in slightly uneven surfaces when these points are reconstructed by COLMAP.

The reconstruction of the filled donut also successfully replicated topology changes, accurately localizing the bitten area, as illustrated in Figure~\ref{fig6}. However, issues arose due to the donut’s reflective, glossy surface, causing misalignment and inaccuracies in depth information, as highlighted in Figure~\ref{fig7}. These artifacts potentially influence volumetric accuracy because the model normalizes the scale of the reconstructed meshes relative to hand size. Nonetheless, the topology itself remained accurate, showcasing our model's potential in handling reflective surface reconstructions, albeit with noted limitations in positional accuracy.

The sandwich reconstruction, depicted in Figure~\ref{fig5}, exhibited more significant qualitative deficiencies. Portions of the original topology were missing, and the reconstructed mesh after biting showed considerable distortion. These shortcomings likely resulted from inadequate feature matches due to insufficient camera viewpoints within the recorded video. Additionally, the inherent homogeneity and symmetry of sandwich-like objects exacerbate the difficulty in accurate feature extraction and matching, leading to incorrect or incomplete 3D shape representations despite appearing satisfactory in certain 2D video frames.

In summary, the qualitative evaluations reveal strong overall alignment and successful topological differentiation for most tested objects. These outcomes underline the feasibility of our method for capturing the progression of food consumption by modeling bites as distinct sequential events. Nevertheless, future studies should further address challenges identified here, particularly alignment and reconstruction errors introduced by reflective surfaces and homogeneous textures, as illustrated by the sandwich and donut examples.

\begin{figure}[]
    \centering
    \includegraphics[scale=0.21]{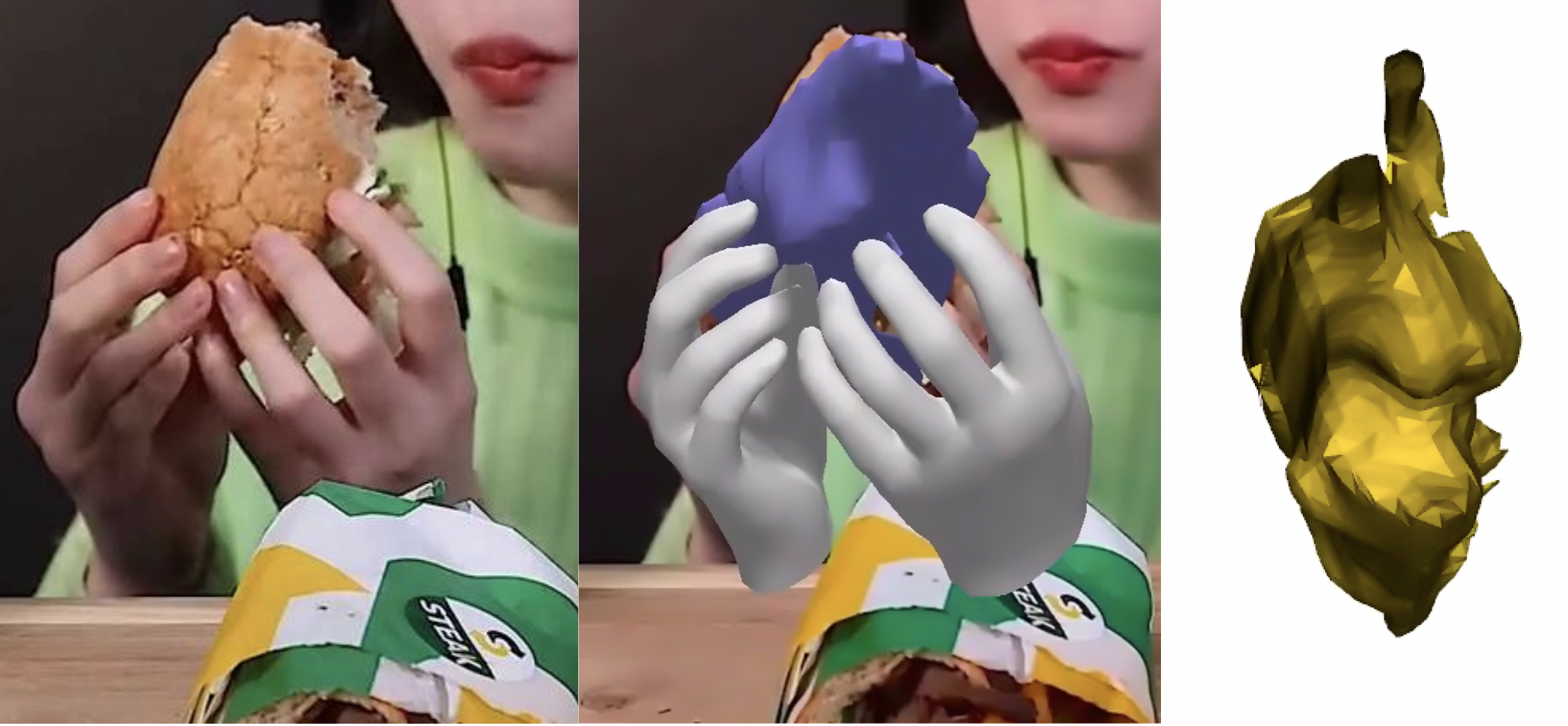}
    \caption{Left: Original RGB Image | Middle: The rendered view of sandwich with hands for out of the wild sequence from YouTube | Right: The 3D mesh of the sandwich}
    \label{fig10}
\end{figure}

For food videos on YouTube, the proposed setup demonstrates notable effectiveness in reconstructing food items, as illustrated in Figure~\ref{fig10}. The reconstructed mesh resembles the sandwich depicted in the original video, successfully capturing its overall shape and key structural details. There are still some discrepancies on the 3D mesh which could be attributed to the lack of different perspective shown in the YouTube video. Despite this promising qualitative result, it is important to acknowledge that no quantitative analysis or precise volume validation was feasible for this particular mesh due to the lack of corresponding ground-truth measurements in the uncontrolled environment. Future work could focus on obtaining appropriate reference data or developing alternative evaluation strategies to systematically validate the model's accuracy and robustness in these real-world scenarios.

\textbf{Quantitative Analysis:}
Based on the meshes reconstructed before and after the detachment of the triceratops' leg, the model estimated a volume decrease of approximately 6.59\%. In comparison, the ground truth measurement using the water displacement method indicated a decrease from around $430 \pm 5 mL$ to $410 \pm 5$ mL. The accuracy of the ground truth measurement itself is constrained primarily by limitations inherent in the experimental apparatus. Specifically, the zip-lock bag used to contain the toy during submersion introduced an estimated uncertainty of $175\pm 5 mL$ due to potential air pockets. As shown in Table~\ref{table1}, the percentage decreases in volume obtained from the ground truth and the model closely align, with only a minor discrepancy of \- 1.348\%. This difference likely arises from variations in the tail structure and differences in mesh resolution, particularly noticeable after topological changes, as highlighted qualitatively in Figure~\ref{fig2}.

For food volume calculations, ground truth values were similarly acquired using the water displacement method. From Table~\ref{table1}, the muffin exhibited a 30.43\% reduction in volume according to the ground truth measurement, whereas the model predicted a slightly higher reduction of 37.70\%, resulting in a difference of $+7.27\%$. The filled donut had a prediction that differed from the ground truth by $+6.6\%$ while the sandwich 3D model differs by $-6.01\%$ in volumetric reduction from the ground truth. Across different food items, the model consistently achieved predictions within roughly $\pm 6\%$ to $\pm 7\%$ of the ground truth measurements. This consistency suggests the proposed model can generalize adequately across various types of food. Although the numerical predictions are not exact, the observed errors remain within an acceptable range. Therefore, the proposed methodology provides sufficiently reliable estimates for dietary intake assessments by measuring the volumetric change of each topological change of the food object.
\section{Conclusion}
\label{conclusion}

In conclusion, this study presented an approach for monitoring food intake through 3D reconstruction using monocular video data. By integrating COLMAP with pose estimation and mesh refinement using the HOLD algorithm, we successfully modeled complex geometry and tracked volume changes. Experiments with toy triceratops and real food items demonstrated the model’s effectiveness in simulating volume reduction, though minor discrepancies were observed due to resolution and occlusion issues.

To further enhance the model’s applicability in automated dietary monitoring, distinguishing between fully intact and partially consumed topology is an important next step. Our proposed approach for automatically separating the unique topologies will be implemented as soon as possible such that the pipeline can make dietary analysis more accurate and adaptable.

{
    \small
    \bibliographystyle{ieeenat_fullname}
    \bibliography{main}
}


\end{document}